%% file: arxiv.tex
\documentclass{article} 
\usepackage{arxiv,times}
\usepackage{booktabs}
\input{math_commands.tex}

\usepackage{graphicx}
\usepackage{hyperref}
\usepackage{url}
\usepackage{algorithm}
\usepackage{algpseudocode}
\usepackage{subcaption}

\title{ORDDAR: Observation-Driven Reasoning for Distortion-Resilient Decision, Action, and Cognitive Recovery}

\author{
Deblina Kar \\
Indian Institute of Technology Kharagpur, India \\
\And
Anant Nawalgaria \\
Google Research, Germany \\
\And
Shyamal Kumar Das Mandal \\
Indian Institute of Technology Kharagpur, India \\
}

\iclrfinalcopy 
\begin{document}

\maketitle

\begin{abstract}
AI agents increasingly perform long-term reasoning, planning, tool use, memory integration, and autonomous decision making, yet erroneous intermediate states can propagate and cause inconsistent decisions and unreliable outputs. Existing reasoning approaches mainly rely on iterative planning, self-reflection, augmented memory, or verification, but rarely localize and selectively repair faulty reasoning. We present
ORDDAR (Observation-Driven Reasoning for Distortion-Resilient Decision, Action, and Cognitive Recovery), a reasoning framework that models reasoning as cognitive state transitions, detects localized distortions, retrieves related reasoning from prior experiences, and repairs only the affected states. ORDDAR therefore performs recovery at the local reasoning-transition level rather than regenerating the complete trajectory. Experiments across mathematical, commonsense, multi-hop, and clinical reasoning benchmarks demonstrate improved reasoning quality, recovery ability, and interpretability over multiple evaluated reasoning baselines.
\end{abstract}

\section{Introduction}

Agentic AI is becoming the next computational paradigm where agents are able to perceive their environments, plan for future actions, use tools from the external world, remember things, and perform elaborate workflows without any human intervention \citep{xu2026theagentcompany}. Nevertheless, robust reasoning still poses a challenge. Mistakes made at some stage of reasoning can cause errors like context drift, inconsistent memory usage, conflicting observations, and bad decisions. Recent benchmarking also indicates that even today's state-of-the-art agentic models fail to reason with complicated instructions and multiple constraints in realistic settings \citep{qi2026agentif}.

Several reasoning paradigms have been proposed to improve autonomous agent reasoning. ReAct integrates reasoning with environment interaction \citep{yao2022react}, Tree of Thoughts (ToT) explores multiple reasoning branches through deliberate search \citep{yao2023tree}, and Reflexion enhances reasoning via self-reflection and episodic memory without model retraining \citep{shinn2023reflexion}. Despite these advances, reasoning stability remains a major challenge, as errors in intermediate reasoning states can propagate through subsequent reasoning trajectories \citep{yeo2025demystifying,gan2025rethinking}. Existing approaches mainly rely on iterative self-correction by repeatedly refining or regenerating reasoning \citep{madaan2023self}, yet their effectiveness varies across models, tasks, and feedback settings \citep{kamoi2024can}. Furthermore, although modern AI agents demonstrate strong planning, tool-use, and memory capabilities, current architectures lack mechanisms for diagnosing localized reasoning failures and selectively repairing only corrupted reasoning states without regenerating the entire reasoning trajectory \citep{wang2024survey}. Consequently, a principled framework for reasoning distortion detection, reasoning-state diagnosis, and selective cognitive recovery remains absent. To address this gap, we propose ORDDAR (Observation-Driven Reasoning for Distortion-Resilient Decision, Action, and Cognitive Recovery).
\begin{figure}[t]
    \centering
    \includegraphics[width=0.5\linewidth]{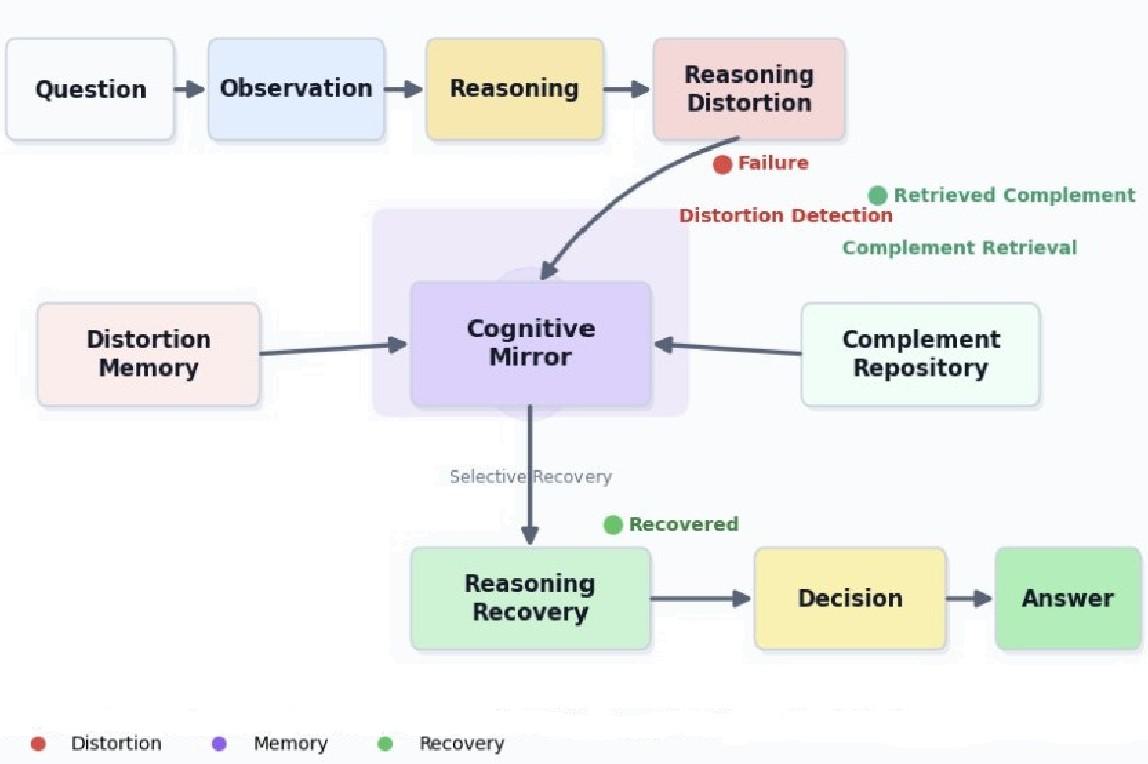}
    \caption{Overview of ORDDAR. The framework detects reasoning distortions, retrieves recovery complements through the Cognitive Mirror, selectively repairs corrupted reasoning states, and predicts the final decision.}
    \label{fig:orddar}
\end{figure}

To address these limitations, we propose ORDDAR (Observation-Driven Reasoning for Distortion-Resilient Decision, Action, and Cognitive Recovery), a novel observation-driven reasoning system which considers reasoning as a series of cognitive state transitions (Fig.~\ref{fig:orddar}). The system identifies local reasoning distortions, analyzes the failure in the reasoning state with a Cognitive Mirror by combining distortion memory with complementary reasoning, and repairs only the faulty reasoning transitions without having to reconstruct the whole reasoning path. The key contributions of our study can be summarized as follows: First, we propose the ORDDAR model that treats reasoning as cognitive state transition and defines localized reasoning distortion. Second, we propose a Cognitive Mirror that integrates distortion memory and complementary reasoning retrieval to identify reasoning failures and recover only corrupted reasoning transition states. Third, we design an end-to-end reasoning pipeline that comprises of reasoning distortion detection, reasoning-state diagnosis, complementary reasoning retrieval, and selective cognitive recovery.

\section{Related Work}
With the fast progress in large language models, agents can now perform reasoning, planning, tool usage, integration of memory, and decision-making in changing environments \citep{wang2024survey,xu2026theagentcompany}. However, recent benchmarking efforts, such as AgentBench, TheAgentCompany, and AGENTIF, reveal that despite significant improvements, state-of-the-art agents still experience difficulties in following instructions, satisfying constraints, and performing consistent decision-making \citep{liu2024agentbench,xu2026theagentcompany,qi2026agentif}. Additionally, frameworks like OpenHands, AutoGen, and MetaGPT augment the capabilities of agents with techniques from software engineering, tool cooperation, and multi-agent coordination \citep{wang2025openhands,wu2024autogen,hong2024metagpt}. Meanwhile, domain-specific benchmarks including MLAgentBench and ScienceAgentBench evaluate autonomous agents on machine learning and scientific discovery tasks, highlighting persistent challenges in reasoning robustness and consistency \citep{huang2023mlagentbench,chen2025scienceagentbench}.

Iterative planning and self-correction have greatly contributed to the improvement of reasoning. ReAct combines reasoning with environment interactions, ToT explores various paths of reasoning, Reflexion and Self-Refine improve reasoning by reflecting and refining it \citep{yao2022react,yao2023tree,shinn2023reflexion,madaan2023self}. However, recently, researchers have found that longer reasoning paths do not always enhance reliability because intermediate reasoning errors tend to propagate and eventually contaminate subsequent reasoning steps \citep{yeo2025demystifying,gan2025rethinking}. In addition, current self-correction methods are extremely sensitive to the models, tasks, and feedback \citep{kamoi2024can}. As a result, most self-correction methods either generate or refine an entire reasoning path instead of fixing corrupted reasoning paths. Distortion of reasoning can be described as the presence of an unusual semantic shift between successive stages of cognition. Reasoning distortion does not preserve semantic integrity while introducing new information, but valid reasoning does so. Hence, semantic disparity is used as an indicator of localized reasoning distortion. The ORDDAR framework specifically identifies reasoning distortions and repairs corrupted reasoning paths instead of generating an entire reasoning path.

\section{Methodology}
A lot of advances have been made by autonomous AI agents in the areas of long-horizon planning, tool utilization, decision making, and lifelong learning, spanning a wide range of areas from embodied agents to web agents, software engineering to autonomous data science \citep{zhou2024webarena,zhang2025deepanalyze,xie2024osworld,park2023generative,luo2025agent,wang2023voyager}. But their reasoning is prone to local errors that cause damage along their reasoning trajectories and compromise the ultimate decisions. In order to compensate for this weakness, ORDDAR proposes a new transition-based cognitive repair approach.

\begin{figure*}[t]
    \centering
    \includegraphics[width=\textwidth]{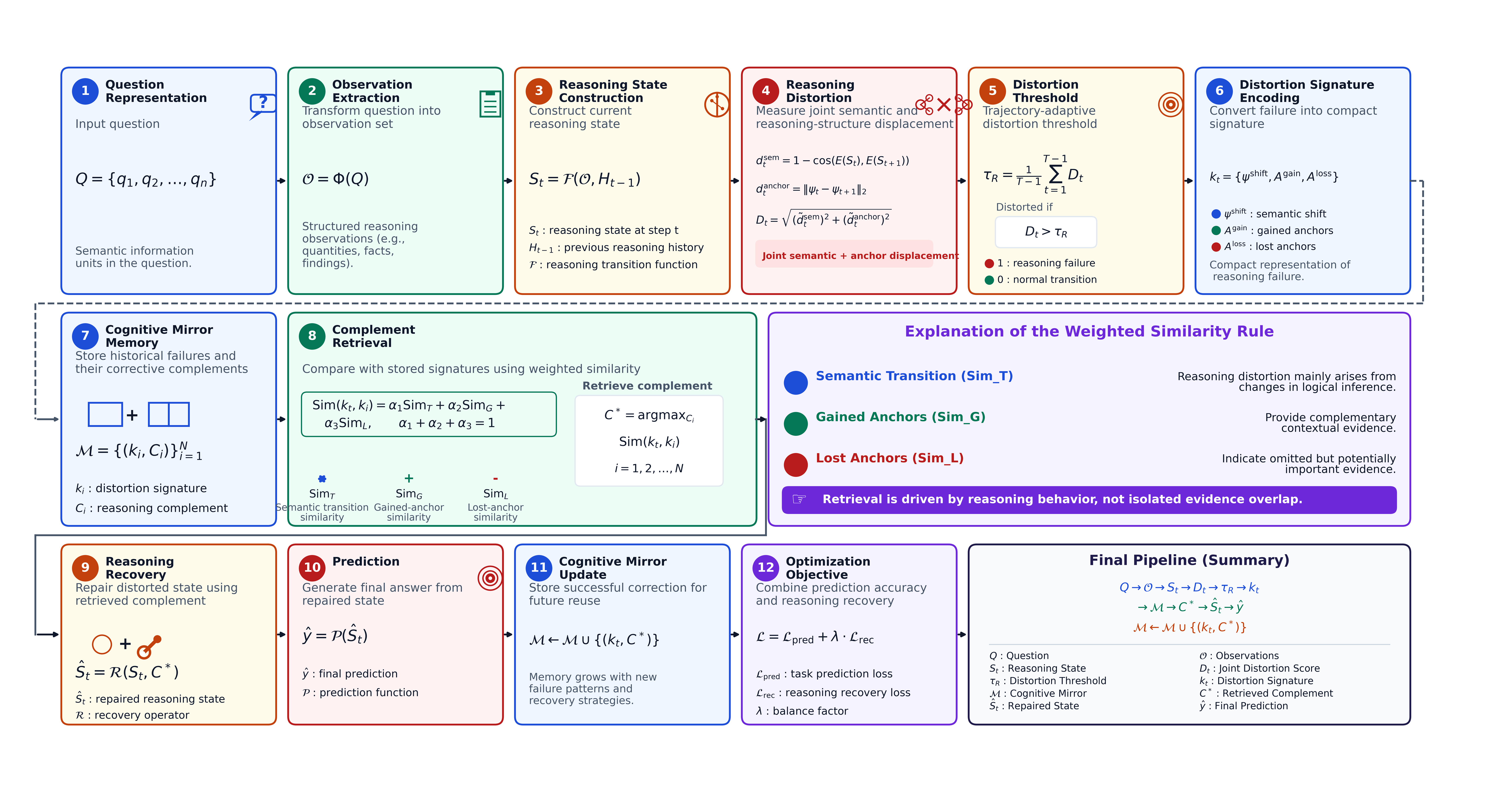}
    \caption{Overall mathematical workflow of ORDDAR. Starting from an input query, ORDDAR constructs observation-driven reasoning states, detects localized reasoning distortions, encodes compact distortion signatures, retrieves the most relevant recovery complement from the Cognitive Mirror using the proposed weighted similarity rule, selectively repairs distorted reasoning states, updates memory with successful recovery patterns, and performs the final prediction.}
    \label{fig:Mathematical}
\end{figure*}

\subsection{Problem Formulation}
Autonomous AI agents solve complex tasks through sequential reasoning, where observations are progressively transformed into cognitive states before decision making. Errors arising during these transitions propagate through subsequent reasoning and degrade final decisions. ORDDAR addresses this problem by recovering localized reasoning errors instead of regenerating the entire reasoning trajectory.

Given an input query $q \in \mathcal{Q}$, an autonomous AI agent generates an output $y \in \mathcal{Y}$ through sequential reasoning. ORDDAR models this process as a sequence of cognitive state transitions, where observations are progressively transformed into intermediate reasoning states. Rather than regenerating the entire reasoning trajectory after failure, ORDDAR identifies and selectively repairs localized reasoning distortions.

The input query is first transformed into a structured observation set
\begin{equation}
O=\phi(q)=\{o_1,o_2,\ldots,o_m\},
\end{equation}
where $\phi(\cdot)$ denotes the observation extraction function and $o_i$ represents the $i$-th reasoning observation extracted from the input. Based on the observation sequence, the reasoning engine constructs a trajectory of cognitive states
\begin{equation}
R=\{s_1,s_2,\ldots,s_T\},
\end{equation}
where each cognitive state is recursively updated as
\begin{equation}
s_t=f(s_{t-1},o_t),
\end{equation}

with $f(\cdot)$ denoting the reasoning transition function. The objective of ORDDAR is to learn a recovery function
\begin{equation}
\mathcal{F}:R \rightarrow \hat{R},
\end{equation}
Where $\hat{R}$ represents the recovered reasoning path. Reasoning is defined as cognitive state changes driven by observations (Eqs.~(2)--(3)). Intermediate reasoning can be captured and detected for distortion recovery by detecting reasoning distortions, acquiring complementary reasoning, and repairing cognitive states for prediction. For localized reasoning analysis, ORDDAR defines a cognitive window $W_t=\{s_t,s_{t+1}\}$ around the reasoning transition currently under inspection. The window moves sequentially along the trajectory, enabling distortion detection and recovery at the level of an individual transition while preserving the remaining reasoning states.

\subsection{Reasoning Distortion Detection and Signature Encoding}
ORDDAR detects localized reasoning failures by measuring the joint displacement of consecutive cognitive states in both semantic embedding space and cognitive-anchor space. Each cognitive state $s_t$ is represented by a semantic embedding $E(s_t)$ and a normalized cognitive-anchor distribution $\psi_t$. For each transition $e_t=(s_t,s_{t+1})$, the semantic transition distance is defined as
\begin{equation}
d_t^{\mathrm{sem}}
=
1-\cos\left(E(s_t),E(s_{t+1})\right),
\end{equation}

where $E(\cdot)$ denotes the semantic embedding produced by a
pretrained sentence encoder. We evaluate three off-the-shelf
sentence encoders, namely MPNet, all-MiniLM-L6-v2, and BAAI/bge-base-en-v1.5, without task-specific fine-tuning. MPNet is adopted as the primary encoder based on its best preliminary
retrieval performance, while MiniLM and BGE are evaluated to assess the robustness of the proposed distortion representation to the choice of semantic embedding space.
\begin{equation}
d_t^{\mathrm{anchor}}
=
\left\|\psi_t-\psi_{t+1}\right\|_2.
\end{equation}

Because these two distances capture complementary aspects of a reasoning transition, they are normalized within the trajectory and combined using a root-sum-of-squares formulation:
\begin{equation}
D_t
=
\sqrt{
\left(\tilde d_t^{\mathrm{sem}}\right)^2
+
\left(\tilde d_t^{\mathrm{anchor}}\right)^2
},
\end{equation}

where $\tilde d_t^{\mathrm{sem}}$ and
$\tilde d_t^{\mathrm{anchor}}$ denote the normalized semantic and anchor-space distances, respectively. The resulting joint distortion score $D_t$ captures changes in both the semantic content and the reasoning structure of consecutive cognitive states. For each reasoning trajectory, we define a trajectory-adaptive distortion threshold as the mean joint distortion score across its consecutive state transitions:

\begin{equation}
\tau_R
=
\frac{1}{T-1}
\sum_{t=1}^{T-1}D_t.
\end{equation}

A transition is considered distorted when
\begin{equation}
z_t=
\begin{cases}
1,&D_t>\tau_R,\\
0,&\text{otherwise}.
\end{cases}
\end{equation}

Here, $\tau_R$ denotes the trajectory-adaptive distortion threshold, and $z_t$ indicates whether the transition is identified as distorted. Only transitions with $z_t=1$ are forwarded for signature encoding. Each detected distortion is encoded as

\begin{equation}
k_t=
\{\psi^{\mathrm{shift}},A^{\mathrm{gain}},A^{\mathrm{loss}}\},
\end{equation}

where $\psi^{\mathrm{shift}}$ denotes the semantic transition shift, $A^{\mathrm{gain}}$ represents newly introduced reasoning anchors, and $A^{\mathrm{loss}}$ denotes discarded reasoning anchors. The Shift--Gain--Loss signature characterizes the nature of the detected distortion and supports recovery-complement retrieval. The resulting signature enables efficient retrieval and cognitive recovery
(Fig.~\ref{fig:Mathematical}). An alternative design is to employ an additional drafter or verifier LLM to assess the correctness of intermediate reasoning states, similar in spirit to verifier-based speculative decoding. We considered this possibility but do not introduce an additional LLM verifier into ORDDAR. Unlike speculative decoding, where a drafter proposes candidate tokens for subsequent verification, ORDDAR operates at the reasoning-state transition level and aims to provide a lightweight, model-independent signal for localizing reasoning distortions. Introducing an additional verifier LLM would add another model-dependent correctness signal and could confound the contribution of the proposed distortion detector. Instead, ORDDAR separates distortion localization from correctness evaluation and evaluates the detector independently using ground-truth reasoning-transition annotations.

\subsection{Cognitive Mirror Memory}
To support efficient reasoning recovery, ORDDAR maintains a \emph{Cognitive Mirror} that stores compact distortion signatures and their corresponding recovery complements, enabling efficient retrieval without preserving complete reasoning trajectories. The Cognitive Mirror is represented as
\begin{equation}
\mathcal{M}
=
\left\{
(k_i,c_i)
\right\}_{i=1}^{N},
\end{equation}
where $k_i$ denotes the encoded distortion signature of the $i$-th reasoning failure and $c_i$ denotes its associated reasoning complement. The Cognitive Mirror stores reusable distortion signatures and recovery complements for efficient indexing and cross-task reuse. Unlike conventional retrieval-augmented reasoning, retrieval in ORDDAR is conditioned on a localized distortion signature and is used specifically to repair the corresponding reasoning transition rather than to augment the entire input context. Each memory entry is created from a detected distorted transition. For a distorted transition $e_t$, ORDDAR encodes its distortion signature $k_t$ and associates it with a recovery complement $c_t$. The signature $k_t$ serves as the retrieval index, while $c_t$ represents the reusable recovery pattern. Thus, the Cognitive Mirror stores compact distortion--recovery pairs rather than complete reasoning trajectories.

\subsection{Complement Retrieval}
At runtime, the current cognitive window is encoded using the same distortion-signature representation. When a distortion is detected, its signature is used to retrieve the most compatible recovery complement from the Cognitive Mirror, which is applied only to the affected reasoning state. For a distortion signature $k_t$, ORDDAR retrieves the most relevant recovery complement from the Cognitive Mirror by computing similarity with stored signatures:
\begin{equation}
S(k_t,k_i)
=
\alpha_1 S_{\mathrm{shift}}
+
\alpha_2 S_{\mathrm{gain}}
+
\alpha_3 S_{\mathrm{loss}},
\qquad
\alpha_1+\alpha_2+\alpha_3=1,\quad
\alpha_j\geq0.
\end{equation}
The weights are constrained to be non-negative and sum to one, where $\alpha_1$, $\alpha_2$, and $\alpha_3$ control the relative contributions of transition shift, gain, and loss similarities respectively. Sensitivity analysis identifies
$(\alpha_1,\alpha_2,\alpha_3)=(0.6,0.2,0.2)$ as the best-performing configuration, achieving the highest average signature matching score. We therefore adopt the 60--20--20 configuration for all subsequent experiments.
\begin{equation}
c^{*}
=
\arg\max_{(k_i,c_i)\in\mathcal{M}}
S(k_t,k_i),
\end{equation}
where $c^{*}$ is the retrieved recovery complement.

\subsection{Localized Reasoning Recovery}
ORDDAR repairs the distorted state using the retrieved complement while preserving the rest of the reasoning trajectory, enabling localized recovery instead of complete regeneration. The restored reasoning state can be obtained as
\begin{equation}
\hat{s}_t
=
\mathcal{R}(s_t,c^{*}),
\end{equation}
where $s_t$ denotes the distorted reasoning state, $c^{*}$ is the retrieved reasoning complement obtained from the Cognitive Mirror, and $\mathcal{R}(\cdot)$ denotes the complement-guided recovery operator.

The recovered reasoning trajectory is 
\begin{equation}
\hat{R}
=
\{s_1,\ldots,\hat{s}_t,\ldots,s_T\},
\end{equation}
where only the distorted reasoning state is replaced while the remaining reasoning trajectory is preserved. This localized recovery avoids unnecessary regeneration of valid reasoning states.

\subsection{Prediction and Cognitive Mirror Update}
The recovered reasoning trajectory is used for final prediction:
\begin{equation}
\hat{y}=P(\hat{R}),
\end{equation}
where $\hat{R}$ denotes the recovered reasoning trajectory and $P(\cdot)$ is the prediction function. Following localized recovery, the recovered trajectory is validated
before the Cognitive Mirror is updated:
\begin{equation}
\mathcal{M}\leftarrow\mathcal{M}\cup\{(k_t,c^{*})\},
\end{equation}
where $k_t$ is the current distortion signature and $c^{*}$ is the retrieved recovery complement. Only validated recovery patterns are eligible for online memory updates. During evaluation, the current test instance is queried against the existing Cognitive Mirror and is not inserted into memory before prediction, preventing evaluation-time contamination. To prevent data contamination, evaluation trajectories are not inserted into the Cognitive Mirror before prediction; only validated recovery patterns are eligible for subsequent memory updates. The complete Cognitive Mirror workflow, including memory construction and indexing, local cognitive-window formation, distortion-signature retrieval, localized reasoning recovery, and validated online memory update, is summarized in Algorithm~\ref{alg:cognitive_mirror}.
\begin{algorithm}[t]
\caption{Cognitive Mirror Construction, Trace-Guided Retrieval, and Recovery}
\label{alg:cognitive_mirror}
\small
\begin{algorithmic}[1]

\Require Validated trajectories $\mathcal{D}$, query $q$,
threshold $\tau_{\mathcal R}$, weights
$\alpha_1,\alpha_2,\alpha_3$
\Ensure Prediction $\hat{y}$

\State Initialize Cognitive Mirror $\mathcal{M}\leftarrow\emptyset$

\Statex \textbf{Memory Construction}
\For{each validated trajectory
$\mathcal{R}_j=\{s_1,\ldots,s_T\}\in\mathcal{D}$}
    \For{$t=1$ to $T-1$}
        \State $W_t\leftarrow\{s_t,s_{t+1}\}$; compute joint distortion $D_t$
        \If{$D_t>\tau_{\mathcal R_j}$}
            \State $k_t\leftarrow
            \{\psi_{\mathrm{shift}},A_{\mathrm{gain}},A_{\mathrm{loss}}\}$
            \State Store validated pair $(k_t,c_t)$ in $\mathcal{M}$
        \EndIf
    \EndFor
\EndFor

\Statex \textbf{Runtime Detection and Recovery}
\State Generate $\mathcal{R}=\{s_1,\ldots,s_T\}$ from $q$

\For{$t=1$ to $T-1$}
    \State $W_t\leftarrow\{s_t,s_{t+1}\}$; compute joint distortion $D_t$
    \If{$D_t>\tau_{\mathcal R}$}
        \State Encode distortion trace
        $k_t\leftarrow
        \{\psi_{\mathrm{shift}},A_{\mathrm{gain}},A_{\mathrm{loss}}\}$
        \State Search $\mathcal{M}$ for a compatible 		distortion trace and retrieve its complement $c^*$ using
		$\alpha_1S_{\mathrm{shift}}+
		\alpha_2S_{\mathrm{gain}}+
		\alpha_3S_{\mathrm{loss}}$
		\If{a compatible trace--complement pair exists}
    			\State $\hat{s}_t\leftarrow R(s_t,c^*)$
    			\State Replace only the affected state
		\Else
    			\State Preserve the current reasoning trajectory
		\EndIf
    \EndIf
\EndFor

\State $\hat{\mathcal R}\leftarrow$ recovered reasoning trajectory
\State $\hat{y}\leftarrow P(\hat{\mathcal R})$

\Statex \textbf{Validated Memory Update for Future Queries}
\If{recovery is validated and memory updating is permitted}
    \State $\mathcal{M}\leftarrow\mathcal{M}\cup\{(k_t,c^*)\}$
\EndIf

\State \Return $\hat{y}$

\end{algorithmic}
\end{algorithm}

The workflow is task-agnostic: although representations and validation may vary across tasks, cognitive-window construction, distortion-trace encoding, complement retrieval, and recovery remain unchanged. The Cognitive Mirror is initialized from validated trajectories; recovery
occurs only when a compatible validated complement exists, otherwise the trajectory is preserved. Successful recovery may update memory for future queries, while evaluation instances are excluded before prediction.

\section{Experimental Setup}
Experiments are conducted on four reasoning benchmarks: GSM8K, HotpotQA, StrategyQA, and CKDQA, covering mathematical, multi-hop, commonsense, and clinical reasoning, respectively. All datasets are publicly available except CKDQA, which was constructed from hospital cases following ethical clearance. For fair comparison, all methods use Qwen2.5-7B-Instruct with its standard tokenizer and chat template, deterministic greedy decoding (\texttt{do\_sample=False}), and a maximum generation length of 512 tokens. Each evaluation uses 100 test instances sampled with random seed 42. For semantic retrieval, ORDDAR evaluates MPNet, all-MiniLM-L6-v2, and BAAI/bge-base-en-v1.5, with MPNet selected based on preliminary retrieval performance; independent ablations evaluate the contributions of the individual ORDDAR components.

\section{Results}

\subsection{Overall Performance}
Comparison of ORDDAR with CoT, Self-Refine, Reflexion, Tree of Thoughts (ToT), and ProCo is provided in Table~\ref{tab:main_results}. ORDDAR achieves the highest reported performance among the compared methods. The difference between the baselines suggests that the reasoning refinement process is highly dependent on task-specific features and intermediate errors. Whereas CoT and Self-Refine depend on a consistent reasoning flow, Reflexion, ToT, and ProCo may still be influenced by intermediate errors or ineffective corrections. ORDDAR identifies reasoning distortions and retrieves complementary reasoning patterns to repair the affected reasoning state.

The higher performance on CKDQA suggests that localized errors in clinical findings or staging criteria can have a significant influence on the final decision. In contrast, HotpotQA features multi-hop evidence dependencies, resulting in more distributed reasoning errors. This indicates that refinement effectiveness depends on both the error type and the nature of the reasoning task. ORDDAR addresses these inconsistencies through localized state detection and recovery.
\begin{table}[t]
\centering
\caption{Performance comparison of CoT, Self-Refine, Reflexion, ToT, ProCo, and ORDDAR. Values denote accuracy (\%) unless otherwise indicated. For HotpotQA, both EM and F1 are reported.}
\label{tab:main_results}
\begin{tabular}{lcccccc}
\toprule
Dataset & CoT & Self-Refine & Reflexion & ToT & ProCo & ORDDAR \\
\midrule
GSM8K & 81 & 87 & 80 & -- & 50 & 92 \\
HotpotQA (EM) & 63 & 66 & 24 & -- & -- & 67 \\
HotpotQA (F1) & 75 & 78 & 30 & -- & -- & 79.22 \\
StrategyQA & 81 & 80 & 60 & 50 & 70 & 86 \\
CKDQA & 67 & 60 & 53.33 & 30 & 25 & 94.38 \\
\bottomrule
\end{tabular}
\end{table}

\begin{table}[t]
\centering
\caption{Comparison of embedding models used for Cognitive Mirror retrieval. Values denote accuracy (\%) unless
otherwise indicated. For HotpotQA, F1 is reported.}
\label{tab:embedding}
\begin{tabular}{lccc}
\toprule
Dataset & MPNet & MiniLM & BGE \\
\midrule
GSM8K & 92 & 91 & 92 \\
HotpotQA (F1) & 79.22 & 79 & 79.1 \\
StrategyQA & 86 & 85 & 85 \\
CKDQA & 94.38 & 87 & 87 \\
\bottomrule
\end{tabular}
\end{table}
\subsection{Ablation Study}
The contributions of the Complement Memory (CM), Distortion Memory (DM), and Recovery Memory (RM) are evaluated through independent ablations (Table~\ref{tab:ablation}). Removing any component consistently degrades performance, confirming their complementary roles. On GSM8K, removing DM causes the largest drop (92\%$\rightarrow$85\%), highlighting the importance of accurate distortion localization. On StrategyQA, removing CM or DM reduces accuracy from 86\% to 81\%, while removing RM yields 83.3\%. Similarly, on HotpotQA, removing CM reduces EM from 67\% to 65\%, DM
reduces F1 from 79.22\% to 79.0\%, and RM produces the lowest EM (62.5\%). The largest degradation occurs on CKDQA, where removing CM, DM, and RM reduces performance from 94.38\% to 81\%, 78.05\%, and 79.79\%, respectively. These results show that complement retrieval, distortion localization, and localized recovery each contribute to ORDDAR's effectiveness.
\begin{table}[t]
\centering
\caption{Ablation study of the Complement (CM), Distortion (DM), and Recovery (RM) mechanisms. Values denote accuracy (\%) unless otherwise indicated. For HotpotQA, both EM and F1 are reported.}
\label{tab:ablation}
\begin{tabular}{lcccc}
\toprule
Dataset & ORDDAR & w/o CM & w/o DM & w/o RM \\
\midrule
GSM8K & 92 & 90 & 85 & 88 \\
HotpotQA (EM) & 67 & 65 & 64 & 62.5 \\
HotpotQA (F1) & 79.22 & 79.0 & 71.0 & 74.03 \\
StrategyQA & 86 & 81 & 81 & 83.3 \\
CKDQA & 94.38 & 81 & 78.05 & 79.79 \\
\bottomrule
\end{tabular}
\end{table}

\subsection{Ground-Truth Validation of Similarity-Based Distortion Detection}
\begin{figure}[!t]
\centering
\begin{minipage}[t]{0.48\columnwidth}
    \centering
    \includegraphics[width=0.7\linewidth]{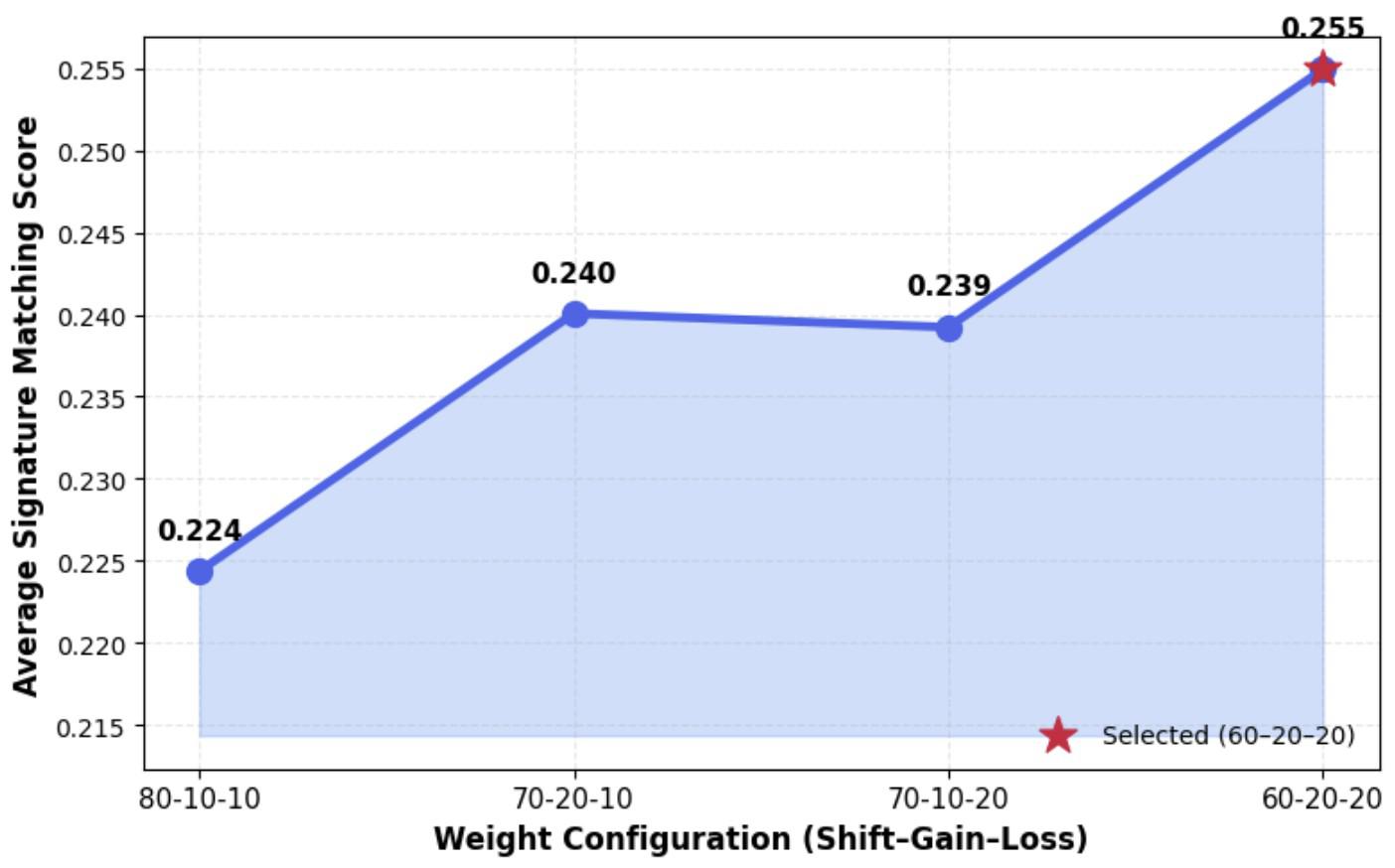}
    \caption*{(a) Validation of the 60--20--20 weighting.}
\end{minipage}
\hfill
\begin{minipage}[t]{0.48\columnwidth}
    \centering
    \includegraphics[width=0.7\linewidth]{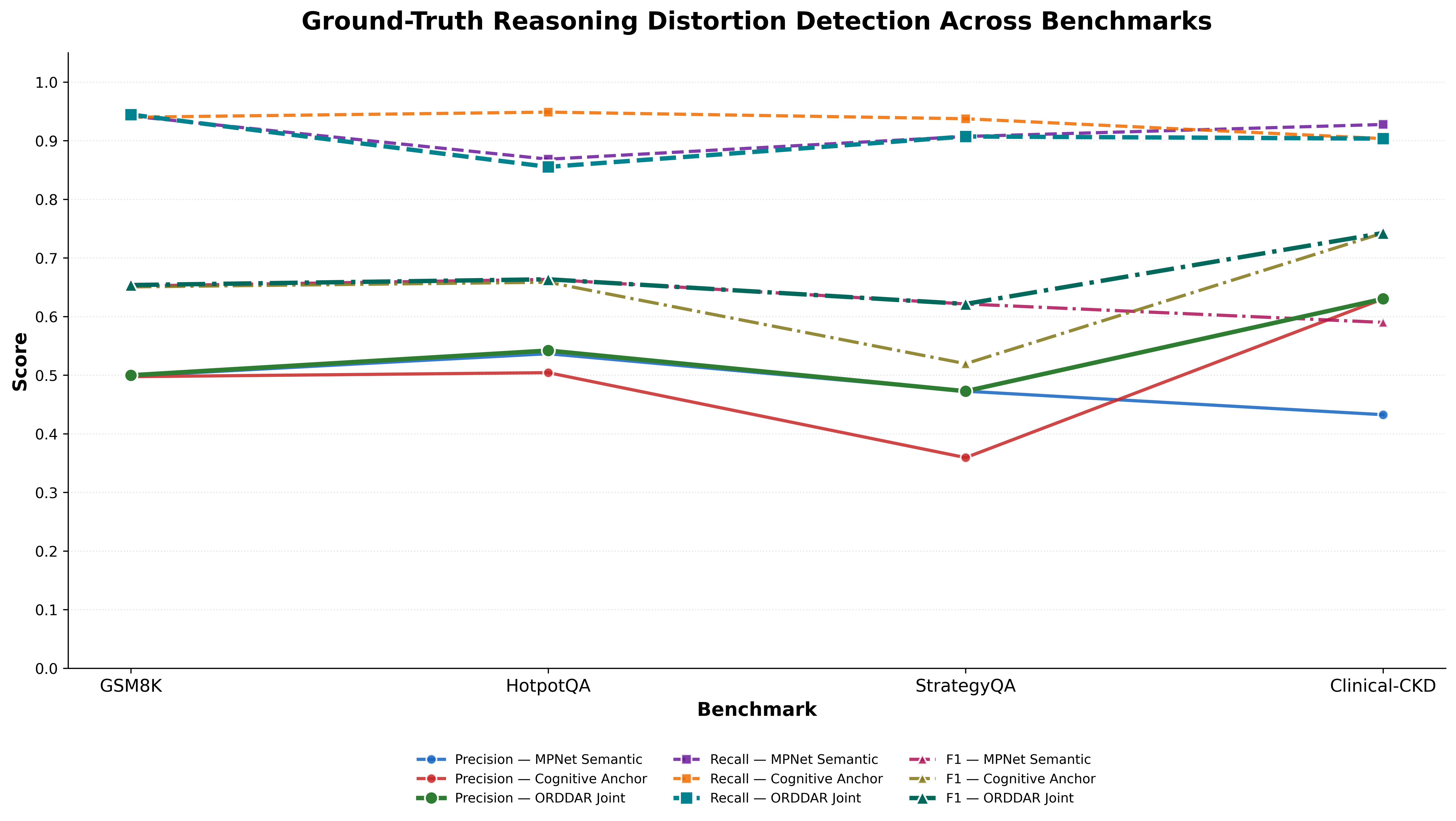}
    \caption*{(b) Ground-truth distortion detection.}
\end{minipage}

\caption{Validation of the proposed distortion representation and its ground-truth detection performance.}
\label{fig:validation_gt}
\end{figure}

To validate the proposed distortion representation, we examine the relative contribution of transition similarity, logical gain, and logical loss. Sensitivity analysis identifies the 60--20--20 weighting as the best-performing configuration, assigning 60\% to transition similarity and
20\% each to gain and loss, as shown in Fig.~\ref{fig:validation_gt}(a). We evaluate embedding-based distortion detection on an independently labeled GT set across GSM8K, HotpotQA, StrategyQA, and Clinical-CKD. Table~\ref{tab:gt_compact} reports precision, recall, and F1 for MPNet Semantic, Cognitive Anchor, and ORDDAR Joint. The cross-benchmark results in Fig.~\ref{fig:validation_gt}(b) show that similarity provides a meaningful distortion signal, with anchor-based representations providing complementary evidence.
\begin{table}[t]
\centering
\caption{Ground-truth validation of reasoning distortion detection. P, R, and F1 denote precision, recall, and F1-score, respectively.}
\label{tab:gt_compact}
\resizebox{\linewidth}{!}{%
\begin{tabular}{l|ccc|ccc|ccc}
\toprule
& \multicolumn{3}{c|}{MPNet Semantic}
& \multicolumn{3}{c|}{Cognitive Anchor}
& \multicolumn{3}{c}{ORDDAR Joint} \\
Benchmark & P & R & F1 & P & R & F1 & P & R & F1 \\
\midrule
GSM8K
& .499 & .942 & .652
& .498 & .940 & .651
& .500 & .945 & .654 \\

HotpotQA
& .537 & .869 & .663
& .504 & .949 & .659
& .542 & .855 & .664 \\

StrategyQA
& .473 & .907 & .622
& .359 & .938 & .520
& .473 & .907 & .622 \\

Clinical-CKD
& .433 & .928 & .590
& .630 & .904 & .743
& .630 & .904 & .743 \\
\bottomrule
\end{tabular}%
}
\end{table}

\subsection{Qualitative Case Study and Retrieval Analysis}

\begin{figure}[!t]
\centering

\begin{minipage}[t]{0.48\columnwidth}
    \centering
    \includegraphics[width=1.2\linewidth]{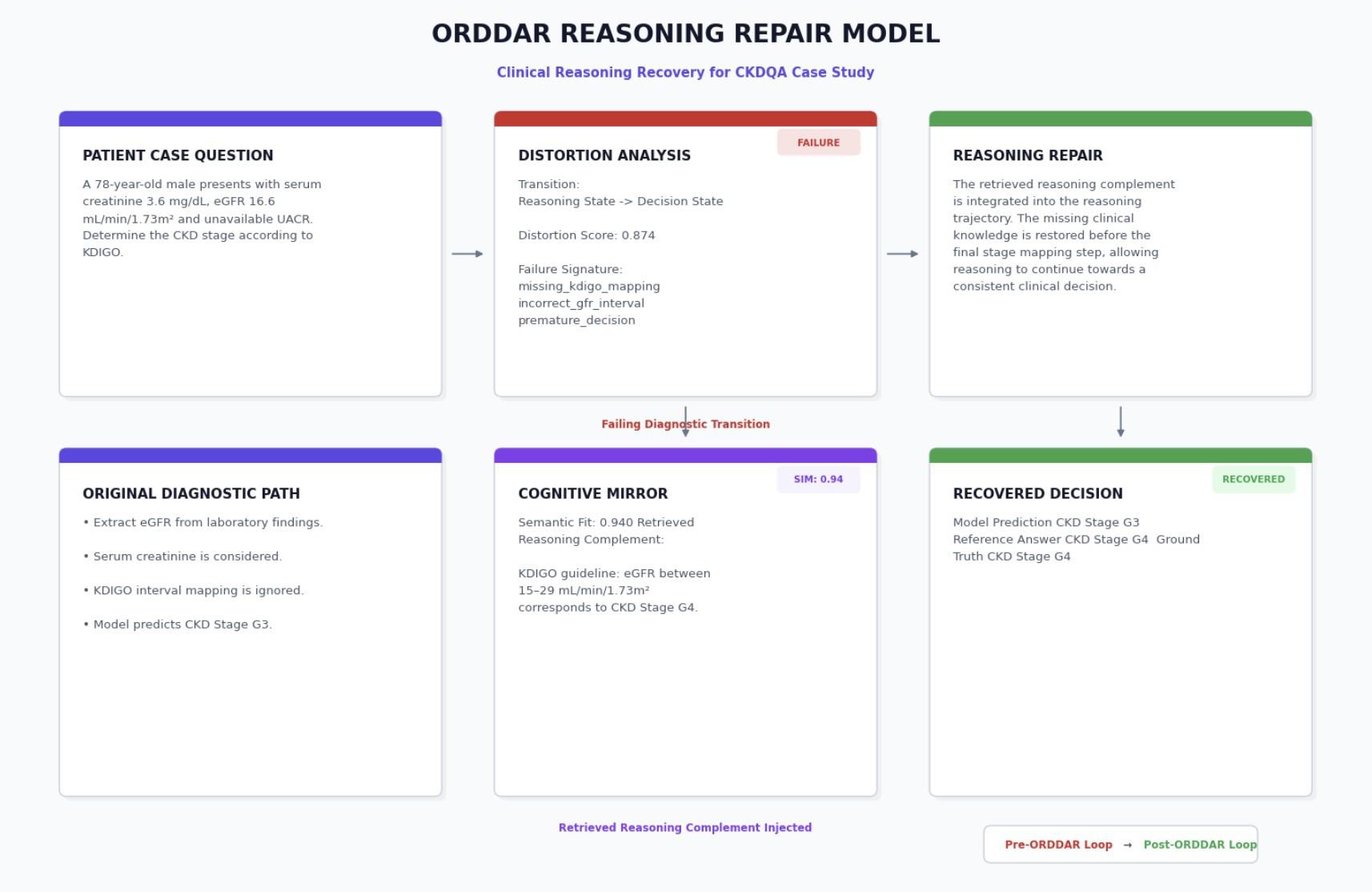}
    \caption*{(a) Clinical-CKD reasoning recovery case.}
\end{minipage}
\hfill
\begin{minipage}[t]{0.48\columnwidth}
    \centering
    \includegraphics[width=0.9\linewidth]{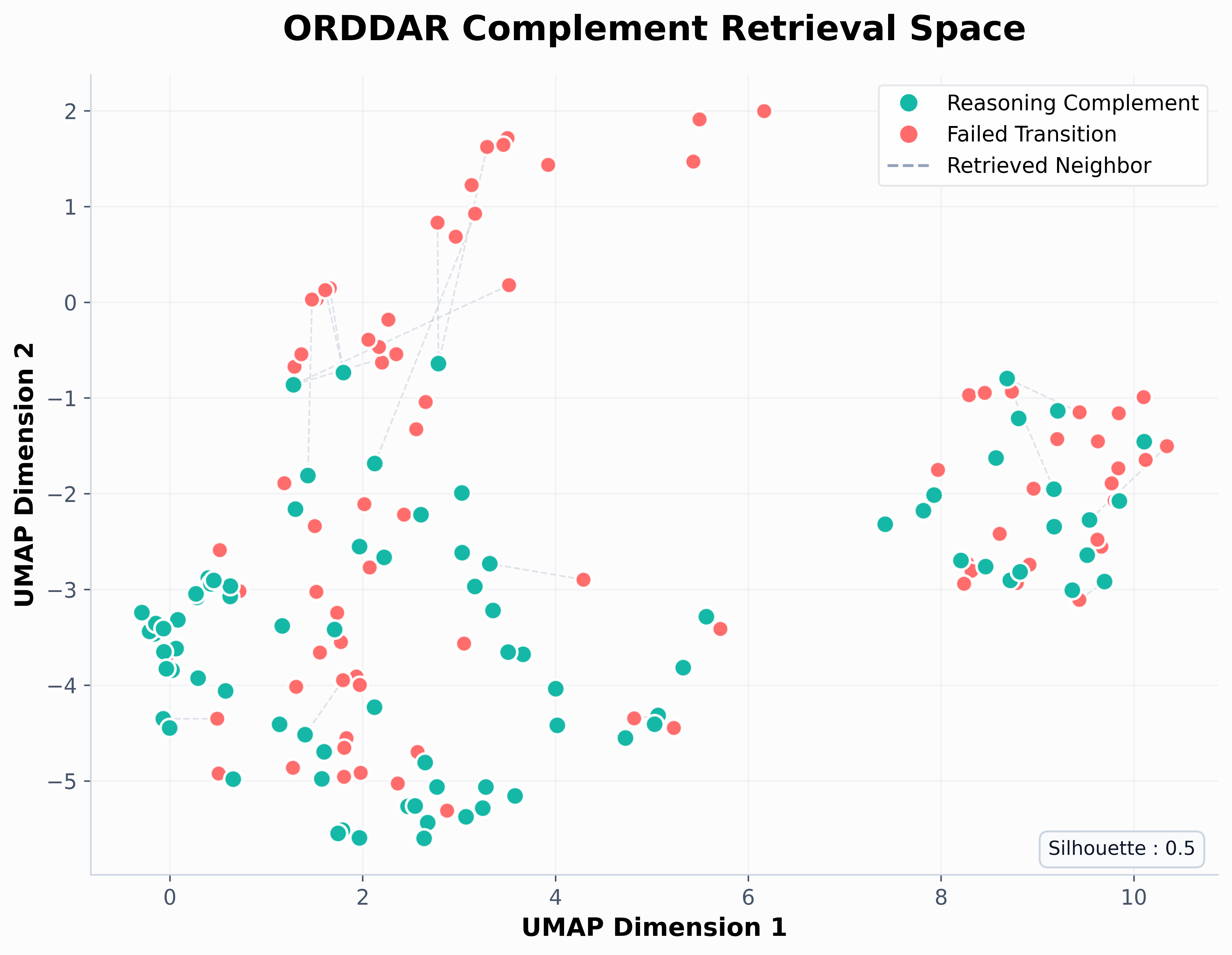}
    \caption*{(b) Cognitive Mirror retrieval space.}
\end{minipage}

\caption{Qualitative analysis of localized reasoning recovery and
Cognitive Mirror retrieval.}
\label{fig:qualitative_analysis}
\end{figure}
A representative Clinical-CKD recovery case is shown in Fig.~\ref{fig:qualitative_analysis}(a). For a patient with creatinine 3.6~mg/dL and eGFR 16.6~mL/min/1.73m$^2$, the initial reasoning incorrectly assigns G3, whereas 15--29 corresponds to G4 under KDIGO \citep{stevens2024kdigo}. ORDDAR localizes the distortion (0.874), retrieves a Cognitive Mirror complement (0.940), and repairs only the erroneous transition, consistent with prior findings on intrinsic self-correction limitations \citep{tyen2024llms,kamoi2024can}. The corresponding UMAP visualization in Fig.~\ref{fig:qualitative_analysis}(b) shows that retrieved complements remain semantically close to detected distortions, suggesting that distortion signatures preserve useful information for recovery-pattern retrieval.

\section{Conclusion}
ORDDAR was introduced as an observation-based framework for detecting localized reasoning distortions and performing cognitive recovery. By representing reasoning as a sequence of cognitive state transitions, ORDDAR detects distorted transitions, retrieves their complements from the Cognitive Mirror, and recovers only the affected states while preserving correct reasoning. Experiments across mathematical, commonsense, multi-hop, and clinical reasoning
consistently demonstrate the superiority of ORDDAR over the evaluated reasoning baselines, particularly on CKDQA. Ablation results confirm that distortion detection, complement retrieval, and recovery each contribute to the final performance. Retrieval analysis further demonstrates that distortion signatures can identify relevant recovery
patterns. Overall, the results show that targeted cognitive recovery can provide an effective alternative to complete reasoning regeneration, while handling higher-order multi-hop dependencies remains an important direction for future research. In the current clinical setting, recovery complements incorporate locally provided KDIGO knowledge. Future work will extend this design to a hybrid recovery scheme combining Cognitive Mirror retrieval with dynamically
retrieved clinical knowledge for broader and more adaptive domain-specific recovery.

\subsection*{AI use statement}

Generative AI tools were used to assist with language
editing and limited code assistance during the preparation of this manuscript. These tools were not used to generate experimental results, figures, perform the reported evaluations, or make final scientific decisions. All AI-assisted outputs were reviewed and verified by the authors. The authors take full responsibility for the final content of this work, including all claims, analyses, results, and artifacts produced with the aid of generative AI.

\subsection*{Ethics statement}

The clinical data used in this study were collected under the appropriate institutional and hospital ethical approvals. The study procedures followed the ethical requirements and guidelines of the participating institution and hospital. Appropriate measures were taken to protect patient privacy and confidentiality, and no personally identifiable information is reported in the manuscript or supplementary materials. The CKDQA dataset is private and is therefore not publicly released.

\subsection*{Reproducibility statement}

We provide the implementation and configuration details for reproducing the ORDDAR framework and the reported evaluation. The source code covers the ORDDAR framework, reasoning distortion detection, Cognitive Mirror retrieval, localized recovery, and the evaluation procedures across GSM8K, HotpotQA, StrategyQA, and CKDQA. The accompanying \texttt{README.txt} and \texttt{requirements.txt} document the required environment, dependencies, dataset access, model configuration, and execution procedure. The implementation and reproducibility materials are provided as anonymous supplementary material; CKDQA requires authorized access to the underlying private clinical data.

\subsubsection*{Author Contributions}
The authors contributed to the conceptualization, methodology, experiments, analysis, and preparation of the manuscript.

\subsubsection*{Acknowledgments}
We sincerely thank the clinicians and doctors who supported this work by providing the clinical data and valuable domain expertise required for the CKDQA experiments. We also gratefully acknowledge the AI research community for valuable knowledge, insights, and discussions on AI agents that contributed to the development of the ideas explored in this work. We further thank everyone who provided helpful discussions and feedback during the development of this research.

\bibliography{iclr2027_conference}
\bibliographystyle{iclr2027_conference}

\appendix

\end{document}

%% file: math_commands.tex
\usepackage{amsmath,amsfonts,bm}

\def\eqref#1{equation~\ref{#1}}

\def\1{\bm{1}}

\DeclareMathAlphabet{\mathsfit}{\encodingdefault}{\sfdefault}{m}{sl}
\SetMathAlphabet{\mathsfit}{bold}{\encodingdefault}{\sfdefault}{bx}{n}

